# Joint PET-MRI Reconstruction with Diffusion Stochastic Differential Model

Taofeng Xie[1,2,3], Zhuoxu Cui[3], Congcong Liu[3], Chen Luo[1], Huayu Wang[1], Yuanzhi Zhang[4], Xuemei Wang[4], Yihang Zhou[3], Qiyu Jin[1], Guoqing Chen[1], Dong Liang[3], Haifeng Wang[3]

[1]Inner Mongolia University, Hohhot, China,
[2]Inner Mongolia Medical University, Hohhot, China,
[3]Shenzhen Institutes of Advanced Technology, Chinese Academy of Sciences, Shenzhen, China,
[4]Inner Mongolia Medical University Affiliated Hospital, Hohhot, China

## Synopsis

**Motivation:** PET suffers from a low signal-to-noise ratio. Meanwhile, the k-space data acquisition process in MRI is time-consuming by PET-MRI systems. **Goals:** We aim to accelerate MRI and improve PET image quality. **Approach:** This paper proposed a novel joint reconstruction model by diffusion stochastic differential equations based on learning the joint probability distribution of PET and MRI. **Results:** Compare the results underscore the qualitative and quantitative improvements our model brings to PET and MRI reconstruction, surpassing the current state-of-the-art methodologies. **Impact: Joint PET-MRI reconstruction is a challenge in the PET-MRI system. This studies focused on the relationship extends beyond edges. In this study, PET is generated from MRI by learning joint probability distribution as the relationship.**

## Introduction

Recent technological advancements have facilitated the fusion of PET and MRI [1]. PET-MRI allows for the simultaneous acquisition of both functional and anatomical data [2][3][4][5]. The harmonized nature of PET and MRI data makes PET-MRI an invaluable tool in various neurological studies [6]. Structural, contrast, and resolution are disparities in PET and MRI due to their fundamentally different imaging mechanisms. Nevertheless, there are statistical correlations between PET and MRI, particularly when PET employs radiotracers like fluorodeoxyglucose (FDG) [7]. PET often feature lower signal-to-noise ratios and restricted spatial resolution in visual imaging due to the inherently stochastic nature of photon emission processes [8][9][10][11]. In contrast, MRI offers high spatial resolution in anatomical imaging. However, the k-space data acquisition process is time-consuming. The common method reduces data acquisition time for upsampling. Undersampling k-space data leads to aliasing in the reconstructed images [12]. The joint reconstruction process capitalizes on the complementary information of MRI and PET. Meanwhile, the score-based diffusion model can learn the joint probability distribution of multi-modality image [13]. This study aims to develop a joint reconstruction model that effectively harnesses the complementary information residing within both modalities, thereby significantly enhancing the quality of their respective reconstructions.

## Method

The joint reconstruction can be described as

$$Y = \mathcal{D}(X) + \epsilon$$

where different modalities can be written in stacked form

$$Y = \begin{pmatrix} f \\ g \end{pmatrix}, \mathcal{D} = \begin{pmatrix} \mathcal{A} \\ \mathcal{F} \end{pmatrix}, X = \begin{pmatrix} u \\ v \end{pmatrix}, \epsilon = \begin{pmatrix} \varsigma \\ \xi \end{pmatrix}$$

Let $u$ and $v$ represent the PET and MRI to be reconstructed, respectively. $f$ corresponds to the observed PET data, and $g$ refers to the observed MRI data. $\mathcal{A}$ signifies the PET forward operator as the discrete Radon transform [14]. $\mathcal{F}$ stands for the MRI forward operator as the discrete Fourier transform. $\epsilon$ represents the measured noise. $\varsigma$ is associated with Gaussian noise, following a normal distribution, i.e., $\varsigma \sim \mathcal{N}(0, I)$, and $\xi$ denotes Poisson noise, i.e., $\xi \sim P(\lambda)$. By reasonable assumptions, the separation of the multi-modality likelihood can be described as

$$p(f, g | u, v) = p(f | u, v) p(g | u, v) = p(f | u) p(g | v)$$

The posterior probability of the PET-MRI image $(u, v)$ for observed PET-MRI data $(f, g)$ can then be simplified as

$$p(u, v | f, g) \propto p(f | u) p(g | v) p(u, v) \tag{1}$$

We further assume a prior of the form $p(u, v) = \exp(-\alpha R(u, v))$. Maximizing (1) is equivalent to minimizing

$$\arg\min_{u,v} \mathcal{J}(u, v)$$
$$= \arg\min_{u,v} \{-\log p(u, v | f, g)\}$$
$$= -\log p(f | u) - \log p(g | v) + \alpha \mathcal{R}(u, v)$$
$$= \sum_{i}^{M} ((\mathcal{A}u)_i - f_i \log(\mathcal{A}u)_i) + \frac{1}{2\sigma^2} \| \mathcal{F}v - g \|_2^2 + \alpha \mathcal{R}(u, v)$$

where $\mathcal{R}(u, v) = \log P(u, v)$. $u, v$ can be reconstructed when $\log P(u, v)$ was obtained by diffusion model. The proposed model as JR-diffusion comprises of the forward diffusion process and the reverse diffusion process. The forward diffusion process, noise is gradually introduced to convert a complex data distribution into a known prior distribution. It is described as

$$(u_{i+1}, v_{i+1}) = (u_i, v_i) + \sigma_{min} \left( \frac{\sigma_{max}}{\sigma_{min}} \right)^i z, \ i = 1, 2, ..., N.$$

Conversely, in the reverse diffusion process, noise is removed to transform the prior distribution back into the original data distribution. It can obtain the reconstruction result

$$(u_i, v_i) = (u_{i+1}, v_{i+1}) - s_\theta((u_{i+1}, v_{i+1}), t) - \mathcal{A}^*\left[\mathcal{I} - \frac{f_{i+1}}{\mathcal{A}u_{i+1}}\right] - \mathcal{F}^*(\mathcal{F}v_{i+1} - g_{i+1}) + z_i.$$

Fig. 1 shows the framework of the JR-diffusion.

## Results

Our training dataset sourced from the Alzheimer's Disease Neuroimaging Initiative (ADNI) [15], consisted of 167 individuals and 5010 image pairs. We conducted a comparative analysis encompassing our method, the conventional Linear Parallel Level Sets (LPLS) method [16], and the supervised deep learning technique known as Joint ISTA-Net [9]. Fig. 2 illustrates the reconstruction outcomes with a Cartesian undersampling factor of 3, while Fig. 3 depicts outcomes with a factor of 5. The results emphasize both qualitative and quantitative enhancements achieved through the implementation of JR-diffusion in PET and MRI reconstruction, surpassing contemporary state-of-the-art methodologies. To substantiate the efficacy of this supplementary information, MRI contrast results for individual and joint reconstructions are presented in Fig. 4, and PET contrast results are displayed in Fig. 5. A comparison between single reconstruction MRI and joint reconstruction MRI reveals that the absence of complementary information significantly compromises the quality of the reconstruction.

## Conclusions

In this paper, we propose JR-diffusion, a novel model for PET and MRI joint reconstruction. It accurately characterizes the joint distribution, serving as a prior for clean image generation from undersampled data. Numerical experiments confirm JR-diffusion's superior performance in joint reconstruction accuracy compared to tradition imaging and supervised deep learning. The accuracy in learning joint probability distributions of PET and MRI primarily drives this improvement.

## Acknowledgments

Taofeng Xie and Zhuo-Xu Cui contributed equally to this work. This work was partially supported by the National Natural Science Foundation of China (61871373, 62271474, 81830056, U1805261, 81729003, 81901736, 12026603, and 81971611),  the National Key R&D Program of China (2023YFB3811400), the Strategic Priority Research Program of Chinese Academy of Sciences (XDB25000000 and XDC07040000), the High-level Talent Program in Pearl River Talent Plan of Guangdong Province (2019QN01Y986), the Key Laboratory for Magnetic Resonance and Multimodality Imaging of Guangdong Province (2020B1212060051), the Science and Technology Plan Program of Guangzhou (202007030002), the Key Field R&D Program of Guangdong Province (2018B030335001), the Shenzhen Science and Technology Program, Grant Award (JCYJ20210324115810030), and the Shenzhen Science and Technology Program (Grant No. KQTD20180413181834876, and KCXF20211020163408012).

**Figures**

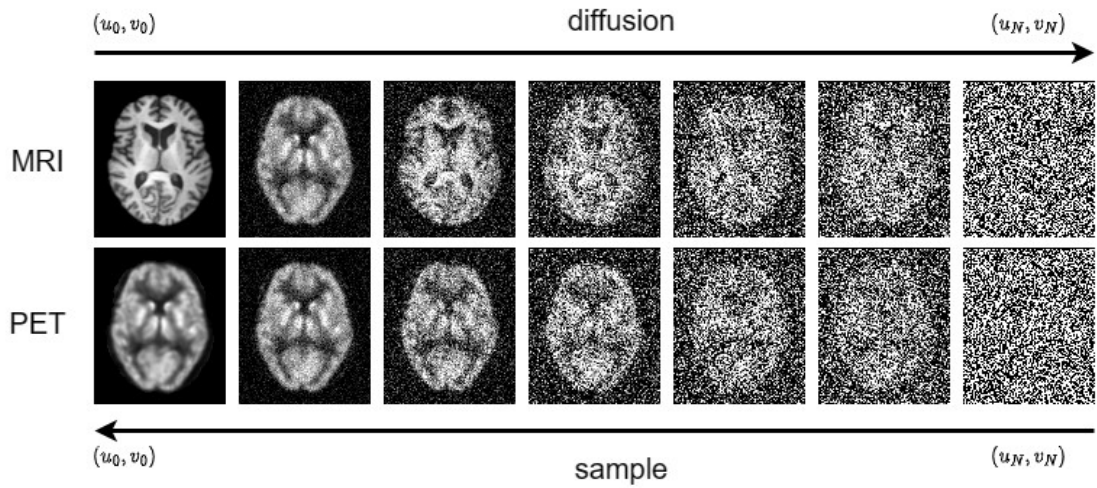

Fig. 1 The schematic of the JR-diffusion model.

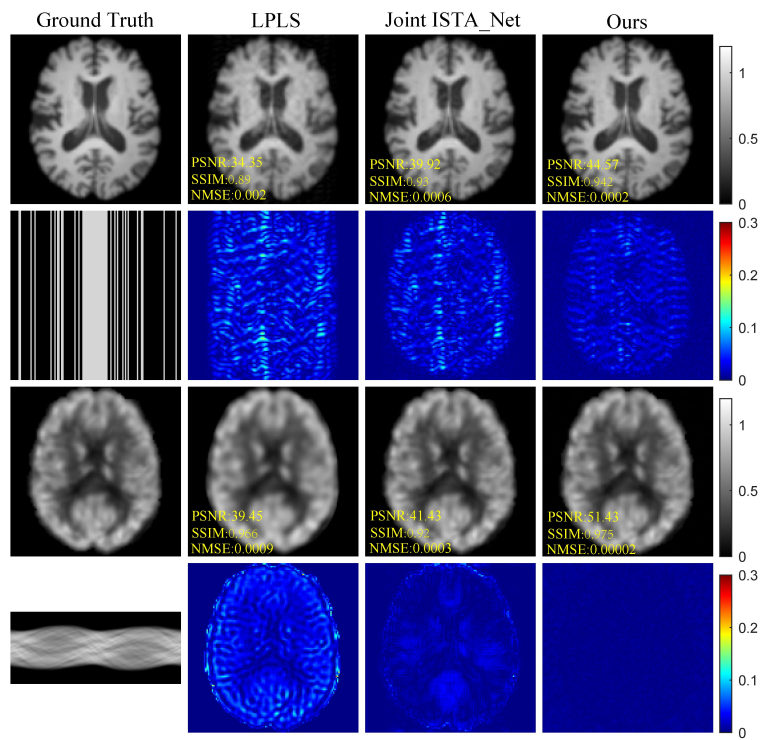

Fig. 2 Joint reconstruction results with cartesian undersampling at 3-fold in k-space and undersampling 128×300 in sinograms.

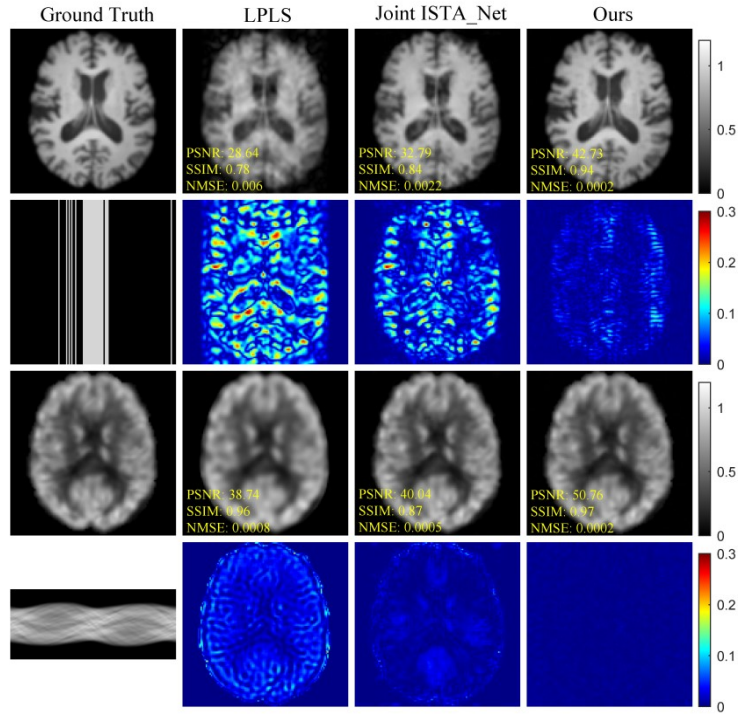

Fig. 3 Joint reconstruction results with cartesian undersampling at 5-fold in k-space and undersampling 128×300 in sinograms.

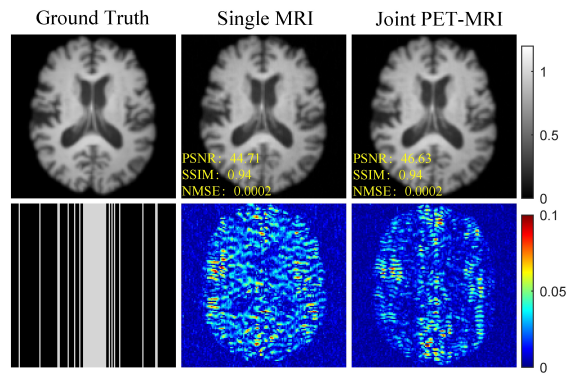

Fig. 4 MRI reconstruction results with cartesian undersampling at 4-fold in k-space for independent reconstruction and joint reconstruction.

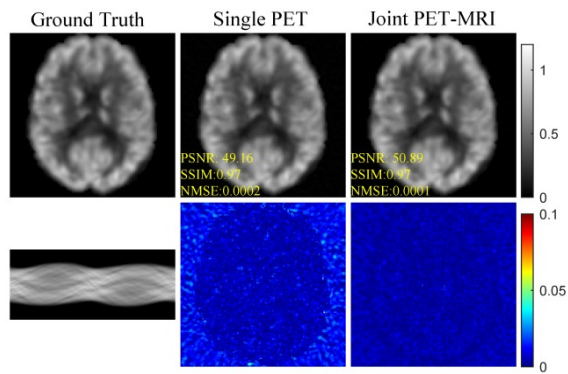

Fig. 5 PET reconstruction results with undersampling 128×300 in sinograms for independent reconstruction and joint reconstruction.